\documentclass[journal,twoside]{IEEEtran}

\usepackage[table,dvipsnames, svgnames, x11names]{xcolor}
\usepackage{graphicx}
\usepackage{amsfonts,amssymb,amsmath,bm,textcomp}
\usepackage{array}
\usepackage[singlelinecheck=false, labelsep=period, font=footnotesize]{caption}
\usepackage{nicematrix}
\usepackage[ruled,vlined]{algorithm2e}
\usepackage{mathrsfs,tabularx,multirow,makecell,setspace,subfig,color,booktabs,multirow,diagbox,xspace}
\usepackage{float}
\usepackage{bm}
\usepackage[american]{babel}
\usepackage[colorlinks, citecolor=black, bookmarks=true, CJKbookmarks=true, urlcolor=black, bookmarksnumbered=true, bookmarksopen=true]{hyperref}
\usepackage{balance}
\usepackage[utf8]{inputenc}
\usepackage{tikz}
\usepackage{bbm}
\usepackage[capitalize]{cleveref}
\usetikzlibrary{spy}

\newcommand{\squishlist}{
	\begin{list}{$\bullet$}
		{ \setlength{\itemsep}{0pt}
			\setlength{\parsep}{2pt}
			\setlength{\topsep}{2pt}
			\setlength{\partopsep}{0pt}
			\setlength{\leftmargin}{1em}
			\setlength{\labelwidth}{1em}
			\setlength{\labelsep}{0.5em} } }
	\newcommand{\squishend}{
\end{list} }

\newcommand{\ldbracket}{{[\kern-0.17em[}}
\newcommand{\rdbracket}{{]\kern-0.17em]}}

\usepackage{fancyhdr}

\begin{document}
	
\title{Deblur-Avatar: Animatable Avatars from Motion-Blurred Monocular Videos}
	
	\author{Xianrui~Luo,~Juewen~Peng,~Zhongang~Cai,~Lei~Yang,~Fan~Yang,~Zhiguo~Cao,~and~Guosheng~Lin

		\thanks{
			\textit{(Corresponding author: Guosheng Lin)}.
			
			Xianrui Luo and Zhiguo Cao are with the School of Artificial Intelligence and Automation, Huazhong University of Science and Technology, Wuhan 430074, China (e-mail: xianruiluo@hust.edu.cn; 
			zgcao@hust.edu.cn).
			
			Xianrui Luo and Zhongang Cai are with the S-Lab for Advanced Intelligence, Nanyang Technological University, Singapore.

            Juewen Peng, Fan Yang, and Guosheng Lin are with the College of Computing and Data Science, Nanyang Technological University, Singapore (e-mail: juewen.peng@ntu.edu.sg; fan007@e.ntu.edu.sg; gslin@ntu.edu.sg).

            Zhongang Cai and Lei Yang are with SenseTime Research, China (e-mail:caizhongang@sensetime.com; yanglei@sensetime.com).
			
			
		}%
	}
	
	\markboth{Manuscript Submitted to IEEE Trans. on Circuit Syst. Video Technol.}%
	{Deblur-Avatar: Animatable Avatars from Motion-Blurred Monocular Videos}
	
	\maketitle
	
    \thispagestyle{fancy}
    \fancyhead{}
    \lhead{}
    \lfoot{}
    \cfoot{\small{Copyright \copyright~2025 IEEE. Personal use is permitted, but republication/redistribution requires IEEE permission.\\See \url{http://www.ieee.org/publications_standards/publications/rights/index.html} for more information.}}
    \rfoot{}


	\begin{abstract}
We introduce a novel framework for modeling high-fidelity, animatable 3D human avatars from motion-blurred monocular video inputs. 
Motion blur is prevalent in real-world dynamic video capture, especially due to human movements in 3D human avatar modeling. 
Existing methods assume sharp inputs, neglecting the motion blur in animatable avatars and failing to address the detail loss introduced by motion.
Our proposed approach integrates a human movement-based motion blur model into 3D Gaussian Splatting. 
By explicitly modeling human motion trajectories during exposure time, we jointly optimize the trajectories and 3D Gaussians to reconstruct sharp, high-quality human avatars. 
We employ a pose-dependent fusion mechanism to distinguish moving body regions, optimizing both blurred and sharp areas effectively.
Extensive experiments on synthetic and real-world datasets demonstrate that our method significantly outperforms existing methods in rendering quality and quantitative metrics, producing sharp avatar reconstructions and enabling real-time rendering under challenging motion blur conditions. Code and models are available at \url{https://github.com/xianrui-luo/deblur_avatar}.

\end{abstract}

\section{Introduction}
\begin{figure*}[!h]
    \centering    
    \includegraphics[width=\linewidth]{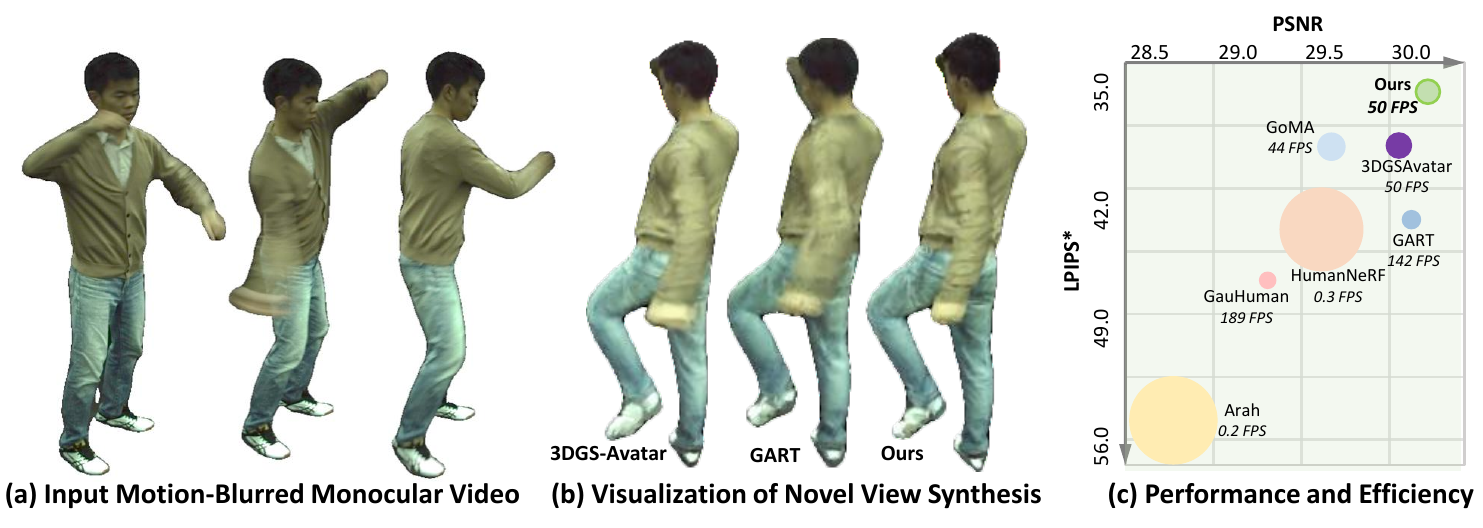}
    \caption{\textbf{We propose the first method to render sharp animatable avatars with a motion-blurred monocular video as input.} 
(b) demonstrates that our method restores sharper details than baselines. (c) LPIPS$^*$=LPIPS$*10^3$. Smaller circles denote higher FPS. 
    }
    \label{fig:figure1}
\end{figure*}
\label{sec:intro}
Reconstructing high-fidelity 3D humans is crucial for applications like AR/VR, film post-production, and gaming. Traditional methods require costly dense multi-view inputs, making them impractical for common use. Reconstructing 3D avatars from sparse-view or monocular inputs is challenging due to underdetermined information and inaccurate body motion estimates. 
To tackle the challenge, implicit neural radiance field (NeRF)~\cite{xu2021h,su2021nerf,noguchi2021neural,lin2022efficient}
is introduced to combine with body articulation~\cite{guo2023vid2avatar,jiang2022neuman,li2022tava,peng2022animatable,liu2021neural,wang2022arah,habermann2023hdhumans} or condition on body-related encodings~\cite{xu2021h,su2021nerf}. However, multi-layer perceptrons (MLPs) in NeRF are computationally expensive, hindering real-time use.
With the advent of 3D Gaussian Splatting (3DGS)~\cite{kerbl20233d}, numerous studies~\cite{hu2024gaussianavatar,jena2023splatarmor,kocabas2024hugs,lei2024gart,li2024animatable,liu2023animatable,liu2024gva,qian20243dgs} integrate 3DGS in human avatar modeling to achieve state-of-the-art rendering quality with fast training and inference. 

Despite the high-quality results, these methods assume the model is trained on videos free from motion blur, where the human remains sharp throughout the entire filming. 
However, as this task requires human movements, encountering motion blur is inevitable in everyday capture. Motion blur occurs when there is significant relative movement between the camera and the human during the exposure time, resulting in a streaking effect as the sensor accumulates light. 
Consequently, motion blur is common in dynamic videos involving humans doing fast movements or in low-light scenes where the exposure time is long so the relative movement is evident.



Current human avatar modeling approaches neglect motion blur, which can severely degrade the quality of reconstructed avatars by obscuring dynamic human motion details.
Several works have proposed to reconstruct static scenes with motion-blurred multi-view inputs, restoring a sharp NeRF or 3DGS by deformable kernels~\cite{ma2022deblur,lee2023dp,peng2024bags} and camera motion model~\cite{wang2023bad,zhao2024bad}. A dynamic pipeline~\cite{sun2024dyblurf} aims to reconstruct dynamic scenes from blurred monocular videos, but its deblurring paradigm is the same as the previous pipelines~\cite{wang2023bad,zhao2024bad}.
These methods perform well on their datasets, however, these methods, including the dynamic one~\cite{sun2024dyblurf}, do not incorporate specialized representations of human motion and would fail if the human movement is more prominent and complex than the camera movement. 



To address motion blur in monocular dynamic human videos, we aim to reconstruct sharp 3D human avatars by explicitly modeling human movements.
We propose an animatable avatar framework to incorporate a human movement-based motion blur model into 3DGS. 
We integrate human motion dynamics into 3DGS training by utilizing an interpolation function to model virtual human motion trajectories over the camera's exposure period. During training, these motion trajectories are optimized jointly with the Gaussians, enabling the reconstruction of sharp human avatars despite motion blur.
Specifically, the trajectory of each motion-blurred image is represented by the initial and final human body poses at the beginning and end of the exposure time, respectively. Given that the exposure time is relatively short, we interpolate between these initial and final poses to determine each intermediate pose along the trajectory. Using this trajectory, we generate a sequence of virtual sharp images by deforming the Gaussians with non-rigid and rigid transforms and projecting the Gaussians onto the image plane. These virtual sharp images are then averaged to synthesize the blurred image, adhering to the physical principles of motion blur. 
Finally, we observe that human movement does not involve the whole body for each frame. Therefore, we propose pose-dependent fusion to effectively distinguish the blurred regions for each pose in a frame, so that our model can focus on modeling motion blur in these blurred areas.

To address the absence of existing human avatar motion deblurring datasets, 
we collect synthetic and real-world datasets and conduct an extensive analysis to demonstrate that our method effectively addresses motion-blurred inputs and renders high-quality sharp avatars, outperforming existing methods.
Additionally, we conduct ablation studies to validate the effectiveness of each component.
In summary, our main contributions are as follows.
\begin{itemize}
    \item[$\bullet$] We present the first framework designed to recover sharp animatable human avatars from monocular videos with motion blur. 
    \item[$\bullet$] We propose human motion trajectory modeling, predicting and interpolating a sequence of human poses to model the motion blur and restore a sharp representation. We also introduce pose-dependent fusion to distinguish moving human body regions for effective training.
    \item[$\bullet$] We evaluate our method on self-collected real and synthetic data, showing substantial quantitative and visual improvements over existing methods.
\end{itemize}







\section{Related Work}
\label{sec:related work}

\noindent\textbf{3D Clothed Human Avatars.}
Traditional methods reconstruct 3D human avatars from dense multi-view stereo~\cite{xu2011video,guo2019relightables}, pixel-aligned features from 3D human ground truth~\cite{alldieck2022photorealistic,saito2019pifu,saito2020pifuhd} or depth cameras~\cite{collet2015high,newcombe2015dynamicfusion,feng2014rapid,xu2019flyfusion}. Explicit representation is introduced to model the human body shapes~\cite{sun2021monocular,kocabas2020vibe} with parametric models~\cite{pavlakos2019expressive,loper2023smpl}, but this representation struggles with topological changes and fails to capture fine details.
With the advent in neural radiance fields~\cite{mildenhall2021nerf} (NeRF), there has been a significant surge in neural rendering for human avatars~\cite{noguchi2021neural,lin2022efficient,jiang2022selfrecon,weng2022humannerf,yu2023monohuman,chen2021animatable} from sparse view videos or a single image~\cite{liao2023high,huang2024tech}. To reconstruct dynamic humans, body model encoding~\cite{su2021nerf,xu2021h} is proposed to enhance generalization, and several works learn deformation field in canonical space~\cite{guo2023vid2avatar,jiang2022neuman,li2022tava,peng2022animatable,liu2021neural,wang2022arah,habermann2023hdhumans} combined with parametric models~\cite{pavlakos2019expressive,loper2023smpl} to model pose and shape, where a latent code~\cite{peng2021neural} for SMPL~\cite{loper2023smpl} vertex is proposed to model appearance, which can be posed by a coordinate-based neural skinning field~\cite{chen2021snarf,mihajlovic2021leap,huang2020arch}.

A disadvantage of NeRF-based methods is their inefficiency in training and rendering. 
3D Gaussian Splatting~\cite{kerbl20233d,guo2024motion} (3DGS) shows superiority in quality and efficiency, so 3DGS is integrated into human avatars~\cite{zielonka2023drivable,hu2024gaussianavatar,jena2023splatarmor,kocabas2024hugs,lei2024gart,li2024animatable,liu2023animatable,liu2024gva,moreau2024human,svitov2024haha,hu2025tgavatar}. KL dispersion~\cite{hu2024gauhuman} and a segmentation plus scaling strategy~\cite{li2024gaussianbody} are used to manipulate point cloud, and kinematic information~\cite{wang2024moss} is used for surface deformation.
Non-rigid deformations~\cite{qian20243dgs,wen2024gomavatar} are integrated into 3DGS to achieve a balance between quality and speed.
Lifted optimization and a novel differentiable shading are introduced ~\cite{wen2024gomavatar,shao2024splattingavatar} to optimize Gaussian parameters while projecting them onto meshes. Contrastive learning~\cite{zhao2024chase} and semantic priors~\cite{zhao2024sg} are used to capture details and maintain consistency.

Although these approaches perform well, they do not consider motion blur in video capture. The inputs are assumed sharp, so current methods deteriorate with motion blur, resulting in low-quality avatars. 
We explore restoring sharp human avatars from motion-blurred monocular video.

\noindent\textbf{Motion Deblurring.}
Traditional single-image deblurring models the blur with normalized gradient sparsity~\cite{krishnan2011blind}, latent structure priors~\cite{bai2019single}, unified probabilistic model~\cite{shan2008high}, and rotational velocity~\cite{whyte10}. Several methods~\cite{nah2017deep, kupyn2018deblurgan, tao2018scale, kupyn2019deblurgan} train neural networks on datasets~\cite{shen2019human}, exploring attention mechanisms~\cite{zamir2022restormer,tsai2022stripformer} and multi-scale pipelines~\cite{ren2023multiscale,dong2023multi}.
On the other hand, video deblurring harnesses temporal information among different frames to ensure consistency. Conventional approaches employ optical flow to model inter-frame relationships~\cite{hyun2015generalized, pan2020cascaded}.
Several approaches stack neighboring frames as the input of neural networks~\cite{su2017deep,zhong2020efficient}. With the new proposed benchmark dataset~\cite{zhang2023mc}, attention mechanisms~\cite{zhang2022spatio,zhang2021multi}, motion magnitude~\cite{wang2022efficient}, and wavelet-based~\cite{pan2023deep} methods are proposed to improve performance and efficiency. While these methods yield excellent results, they assume the blur is predominantly from camera movement, and the method performs poorly on blur from human motion. 
Furthermore, these methods operate in image space and fail to capture scene geometry in $3$D space. 

Recently NeRF has demonstrated impressive results in novel view synthesis, and many methods have extended NeRF from sharp static scenes to motion-blurred static~\cite{ma2022deblur,lee2023dp,wang2023bad,lee2023exblurf} and dynamic scenes~\cite{sun2024dyblurf}. 
However, the static pipelines are not fit for dynamic motion, and the dynamic pipeline~\cite{sun2024dyblurf} is designed for the entire scene, so only the general human motion of the scene can be deblurred. The method lacks specialized human motion representation and would fail if the human movement is more prominent and complex than the camera movement.
To improve efficiency, 3DGS is also introduced to model the blur with the covariance of 3D Gaussian~\cite{lee2024deblurring,darmon2024robust,peng2024bags} and camera motions~\cite{chen2024deblur,zhao2024bad,lee2024crim}. Current methods, including the dynamic deblurring pipeline, are designed specifically for motion blur from camera motions instead of human motions, and would not suffice for our aim: restoring from motion blur due to human movements in human avatar modeling.

\section{Method}
\label{sec:method}
\begin{figure*}[!h]
    \centering    
    \includegraphics[width=\linewidth]{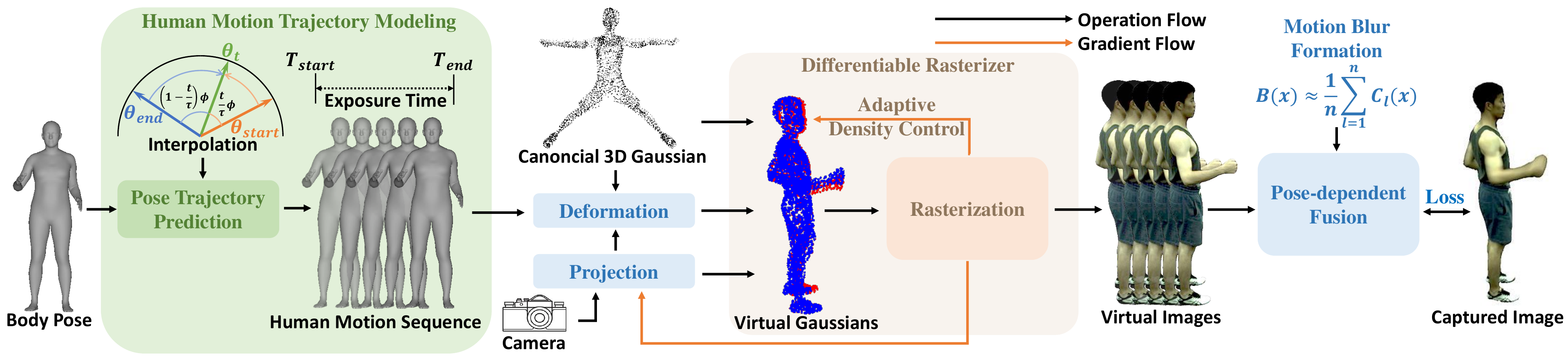}
    \caption{\textbf{Pipeline of our method.} For each frame with its corresponding body pose as input, we introduce human motion trajectory modeling to predict human poses $\theta_{start}$ and $\theta_{end}$ at the start $T_{start}$ and end $T_{end}$ of the exposure period of the frame. Then we use interpolation to obtain a virtual human pose sequence. 
    With the pose of each virtual timestamp, we perform non-rigid deformation and rigid transformation to the 3D Gaussian initialized in canonical space. 
    The observation space virtual 3D Gaussians and their colors are accumulated through differentiable rasterization to render sharp virtual images. These images are averaged as \cref{eq:blur_process} to produce the blurred result. The blurred image is then blended with the virtual sharp result by pose-dependent fusion to render the final output, supervised by the captured input RGB image.
    During testing, Gaussians can be rendered directly to produce a sharp image without motion trajectory modeling and pose-dependent fusion, enabling fast rendering.
    }
    \label{fig:pipeline}
\end{figure*}

We propose a human avatar modeling pipeline to recover sharp novel views and poses from a motion-blurred monocular video. 
In the following, we first revisit the concept of 3DGS and the principle of motion blur. 
In section~\ref{sec:method_3}, we propose human motion trajectory modeling to account for human movements in the exposure period. 
We show how to perform rigid and non-rigid deformation from the canonical space in section~\ref{sec:method_4}.
Then we apply pose-dependent fusion to predict the blurred region of each view, blending the rendered blur region and the rendered sharp region to get optimal results in section~\ref{sec:method_5}. 

\subsection{Preliminary: 3D Gaussian Splatting}
3DGS~\cite{kerbl20233d} represents the 3D scene with a set of discrete 3D Gaussians $G$, parameterized by its mean $\mu \in \mathbb{R}^3$, covariance $\Sigma \in \mathbb{R}^{3 \times 3}$, opacity $\alpha \in \mathbb{R}$, and view-dependent color $c \in \mathbb{R}^3$ represented by spherical harmonics coefficients $f$. The distribution of each scaled Gaussian is defined as:

\begin{equation}
G(\mathbf{x}) = e^{-\frac{1}{2} (\mathbf{x} - \mu)^\top \Sigma^{-1} (\mathbf{x} - \mu)}\,,
\end{equation}

To ensure that the 3D covariance $\Sigma$ remains in its positive semi-definiteness, and to reduce the optimization difficulty, 3DGS represents $\Sigma$ using a scale $S \in \mathbb{R}^3$ and a rotation matrix $R \in \mathbb{R}^{3 \times 3}$ stored by a quaternion $q \in \mathbb{R}^4$:

\begin{equation}
\Sigma = RS S^\top R^\top\,,
\end{equation}

To enable differentiable Gaussian rasterization, the 3D Gaussians are projected to 2D image plane from a given camera pose $W \in \mathbb{R}^{3 \times 3}$ and the Jacobian of the affine approximation of the projective transformation $J \in \mathbb{R}^{2 \times 3}$:

\begin{equation}
\Sigma' = J W \Sigma W^\top J^\top\,,
\end{equation}

where $\Sigma' \in \mathbb{R}^{2 \times 2}$ is the 2D covariance matrix.
The pixel color is thus rendered by rasterizing these $N$ sorted 2D Gaussians according to their depths:

\begin{equation}\label{eq:gs}
\mathbf{C} = \sum_{i=1}^{N} c_i \alpha_i \prod_{j=1}^{i-1} (1 - \alpha_j)\,,
\end{equation}

where $c_i$ is the view-dependent color, and $\alpha_i$ is the learned opacity value $\alpha$ weighted by probability density of $i$-th projected 2D Gaussian at the target pixel location.
3D Gaussians are initialized from Structure-from-Motion~\cite{schoenberger2016sfm} (SfM) or random sampling. 
Adaptive control is then applied to improve rendering quality through split, clone, and prune. Gaussians with large position gradients are split if their size exceeds a threshold or cloned if below it. Gaussians with small $\alpha$ values or overly large scales are pruned.




\subsection{Preliminary: Motion Blur Formation}
\label{sec:method_2}
The physical process of motion blur formation comprises the collection of photons during a certain exposure time $\tau$ due to camera or object motion. The final blurred image can be represented by fusing a sequence of virtual images:
\begin{equation}
    \mathbf{B}(x) = \gamma \int_0^{\tau} \mathbf{C}_t(x) dt\, ,
\end{equation}
where $\mathbf{C}_t(x)$ is the sharp image captured at time $t$, $x$ is the pixel location of the image, $\mathbf{B}(x)$ is the blurred image, and $\gamma$ is a normalization factor. We discretize the exposure period into $n$ timestamps for approximation and the captured result is denoted as the average of $n$ sharp images: 
\begin{equation}
\label{eq:blur_process}
    \mathbf{B}(x) \approx \frac{1}{n} \sum_{l=1}^{n} \mathbf{C}_l(x)\, .
\end{equation}

As shown in \cref{eq:blur_process}, images from each timestamp are required to render the final blurred result. Our task focuses on human movements, therefore, it is important to model human movements during the exposure time. 



\subsection{Human Motion Trajectory Modeling}\label{sec:method_3}
Following previous works, the inputs are a monocular video with motion blur, calibrated camera parameters, SMPL parameters, and foreground segmentation masks. The training initiates with randomly sampling points on the canonical SMPL~\cite{loper2023smpl} mesh surface. 
SMPL is a parametric human model that defines $\beta$ and $\theta$ to control body shapes and poses. Linear Blend Skinning (LBS) algorithm is applied in SMPL to transform a point $p_c$ from a canonical space to an observation space point $p_o$:
\begin{equation}
    p_o = \sum_{k=1}^K w_k \mathbf{T}_kp_c\,,
\end{equation}
where $K$ is the number of joints, 
$\mathbf{T}_k$ is the $4$x$4$ rotation-translation matrix calculated from local joint positions $\mathbf{J}= \left\{J_1,...,J_K\right\}$, shape $\beta$ and pose $\theta$. 


As human poses are stored in SMPL, we propose to model human motion trajectory from SMPL parameters, specifically the human pose $\theta$. For avatar modeling with motion blur, the blur results from human movement in the exposure period. Therefore,   
based on \cref{eq:blur_process}, a straightforward approach to address motion-blurred inputs $\mathbf{B}_{gt}$ is to reconstruct each virtual sharp image $\mathbf{C}_l$, serving as the dependent variable of the motion-blurred output. 
Given that a sharp image $\mathbf{C}_l$ can be rendered from a specified human pose $\theta_l$ within the framework of 3DGS avatar pipeline~\cite{qian20243dgs,hu2024gauhuman}, it is viable to establish correspondence between virtual human poses and rendered virtual sharp images. 
Since we aim to render a sequence of virtual sharp images to depict human movements, 
the corresponding virtual poses of each latent sharp image within the exposure time $\tau$ can be acquired from employing a human motion trajectory represented through Spherical Linear Interpolation~\cite{shoemake1985animating} (Slerp) between two SMPL poses in quaternion, one at the beginning of the exposure $\theta_{\text{start}}$ and the other at the end $\theta_{\text{end}}$. The virtual human pose at time $t \in [0, \tau]$ can thus be expressed as follows:

\begin{equation}
    \theta_t =\frac{\sin(\frac{t}{\tau} \phi)}{\sin\phi} \theta_{\text{start}} + \frac{\sin((1-\frac{t}{\tau}) \phi)}{\sin\phi} \theta_{\text{end}}\,,
\end{equation}
where $\tau$ represents the exposure time. To align with \cref{eq:blur_process}, $\frac{t}{\tau}$ can be further denoted as $\frac{l}{n-1}$ for the $l$-th latent sharp image $\mathbf{C}_l$ corresponding with pose $\theta_l$. $\phi$ is the angle between $\theta_{\text{start}}$ and $\theta_{\text{end}}$, obtained as:
\begin{equation}
    \phi = \arccos(\theta_{\text{start}} \cdot \theta_{\text{end}})\,,
\end{equation}
We aim to estimate both $\theta_{\text{start}}$ and $\theta_{\text{end}}$ for each frame, along with the learnable parameters of Gaussians $G$.
To predict both poses from a single input, we initialize $\theta_{\text{start}}$ and $\theta_{\text{end}}$ with the input body pose parameters. The learnable poses are used for subsequent rigid transformation, allowing direct optimization through backpropagation.
For very small angles (i.e., when $\sin(\phi)$ is close to zero), we assume the corresponding frame has very little motion blur and linear interpolation is used as an approximation. 
We also predict the sequences of per-frame translations with direct linear interpolation and apply Slerp to the global orientation of the body to get a virtual sequence.
We set the virtual frame number $n=5$ as the number of images during an exposure time, balancing training efficiency and rendering quality.

\subsection{Human Deformation and Color Prediction}\label{sec:method_4}
Human deformation can be decomposed into skeleton-based rigid movements and pose-based cloth deformation~\cite{qian20243dgs}. We first apply non-rigid deformation to canonical Gaussians $G_c$, using the canonical positions $x_c$ and the pose-based latent code generated from pose encoder~\cite{mihajlovic2021leap} as inputs. The non-rigid deformation MLP \textbf{$f_{\theta_{\text{nr}}}$} predicts the non-rigid parameters as:

\begin{equation}\label{eq:nonrigid}
(\Delta x, \Delta s, \Delta r) = f_{\theta_{\text{nr}}} (x_{c}; l_{\text{pose}})\,,
\end{equation}
where $\Delta x$ indicates the offset of 3D position of Gaussian, $\Delta s$ is the offset factor of scale, $\Delta r$ is the scaling factor of rotation. Then the Gaussian at position $x_o$ is deformed as:
\begin{equation}
    x_{\text{nr}}=x_c+\Delta x\,, 
\end{equation}
\begin{equation}
    s_{\text{nr}}=s \cdot exp(\Delta s) \,,
\end{equation}
\begin{equation}\label{eq:rotation}
    r_{\text{nr}}=r_c\cdot[1,\Delta r_1,\Delta r_2,\Delta r_3]\,,
\end{equation}
where the dot product in \cref{eq:rotation} is quaternion multiplication, and $[1,0,0,0]$ is denoted as identity rotation.

Now we have the non-rigid-deformed canonical space $G_{\text{nr}}$ and a sequence of virtual human poses, we deform the 3D Gaussians $G_{\text{nr}}$ to observation space with rigid transformation.
For each pose and shape from one virtual frame, we calculate the rotation-translation body transform matrix $\mathbf{T}_k$ from the SMPL parameters the same as prior works~\cite{qian20243dgs}, then a skinning MLP $f_{\theta_{\text{rigid}}}$ predicts the skinning weights. The transformed position $x_{o}$ and the rotation matrix of 3D Gaussians $R_{{o}}$ are shown as:
\begin{equation}
\mathbf{T} = \sum_{k=1}^K f_{\theta_{\text{rigid}}} (x_{\text{nr}})_k \mathbf{T}_k\,,
\end{equation}
\begin{equation}
x_{o} = \mathbf{T} x_{\text{nr}} \,,\
\mathbf{R}_{{o}} = \mathbf{T}_{1:3, 1:3} \mathbf{R}_{\text{nr}}\,,
\end{equation}
the rotation matrix $\mathbf{R}_{\text{nr}}$ is transformed from quaternion $r_{\text{nr}}$. 

We do not apply spherical harmonics coefficients in regular 3DGS to predict the color of each Gaussian. We use a very small MLP with one $64$-dimension hidden layer as previous work~\cite{qian20243dgs} to model the appearance. 


\subsection{Pose-dependent Fusion}\label{sec:method_5}
We observe that a human cannot move every part of their body all at once, so for each blurred frame, there are always sharp regions on the body. 
The sharp regions do not require human motion trajectory modeling and \cref{eq:blur_process} to render. As motion trajectory modeling is prone to reconstruct the region with dynamic priors,
we propose to estimate a per-pixel fusion mask $\mathbf{M}(x, j)$, where $x$ is the same as ~\cref{eq:blur_process}, $j$ is the $j\text{-th}$ frame from original inputs. 
We equip a small MLP $f_{\theta_{\text{fuse}}}$ with the inputs consisting of a latent pose code $l_{\text{pose}}$ from \cref{eq:nonrigid}, a view embedding $l_j$, a position encoding $l_x$, and color embedding $l_{{\text{rgb}}}$ extracted from rendered virtual image $\mathbf{C}(x, j)$. The fusion mask is predicted as:

\begin{equation}
   \mathbf{M}(x, j)=f_{\theta_{\text{fuse}}}(l_{\text{pose}},l_{\text{j}},l_{\text{x}},l_{{\text{rgb}}})\,,
\end{equation}\label{eq:fusion}

where this mask is used to blend the virtual sharp image $\mathbf{C}(x, j)$ with the blurred $\mathbf{B}(x, j)$ (stacked from \cref{eq:blur_process}):
\begin{equation}
\mathbf{C}_{\text{out}}(x, j) = (\mathbf{1}- \mathbf{M}(x, j)) \mathbf{C}(x, j) + \mathbf{M}(x, j) \mathbf{B}(x, j)\,.
\end{equation}
This mask ensures that pixels are learned appropriately based on blur amount. If a pixel is blurred, $\mathbf{M}(x, j)$ will be large, assigning greater weight to the blurred image $\mathbf{B}(x, j)$. Conversely, if a pixel is clear, $\mathbf{M}(x, j)$ will be small, favoring the supposedly sharp virtual image $\mathbf{C}(x, j)$. This sharp virtual result is regarded as the final rendered result. This fusion module helps the network to optimize sharp and blurred regions separately, achieving effective training. This module and the human motion trajectory modeling module are not used for rendering, as these modules are meant for training with the motion-blurred inputs, and we only need the sharp output for inference.

\subsection{Optimization}\label{sec:method_6}

We learn the Gaussians and the human motion trajectories for each frame from the loss functions, consisting of an RGB loss $L_{\text{rgb}}$ and perceptual loss $L_{\text{percept}}$ for $\mathbf{C}_{\text{out}}(x, j)$ and $\mathbf{B}_{\text{gt}}$, a mask loss $L_{\text{mask}}$ for foreground human. 
Skinning weight regularization $L_{\text{skin}}$ and isometric regularization $L_{\text{isopos}}, L_{\text{isocov}}$~\cite{qian20243dgs} are implemented to alleviate noisy deformation. 
The isometric regularization losses regularize both position and covariance $L_{\text{isopos}}$ and $L_{\text{isocov}}$. 
For the full loss function $L_{\text{full}}= L_{\text{rgb}} + \lambda_1 L_{\text{percept}}+\lambda_2 L_{\text{mask}}  + \lambda_3 L_{\text{skin}} \\
+ \lambda_4 L_{\text{isopos}} + \lambda_5 L_{\text{isocov}}$,
$\lambda_1=0.01$, $\lambda_2=0.1$, $\lambda_3$ is set to 10 for initialization and decrease in training. $\lambda_4=1$, $\lambda_5=100$.



\noindent\textbf{Implementations.}
Our model is trained for $15k$ iterations on the ZJU-MoCap-Blur dataset and the Real-Human-Blur dataset in $40$ minutes on a single NVIDIA RTX 3090 GPU. We employ the Adam optimizer~\cite{kingma2014adam} to optimize our model and the per-frame latent codes. The learning rate of 3D Gaussians is unaltered as the original 3DGS implementation~\cite{kerbl20233d}. We initialize the canonical 3D Gaussians with N = $50k$ random samples on the SMPL mesh surface in canonical pose. During optimization, we follow the original 3DGS pipeline~\cite{kerbl20233d} to densify and prune the 3D Gaussians, using the view-space positional gradients to determine the densification regions. 
To alleviate overfitting in color prediction, inverse rigid transformation is implemented in the same way as ~\cite{qian20243dgs} to canonicalize the viewing direction. Then the spherical harmonics basis of canonicalized viewing direction and pose-dependent feature from non-rigid MLP are used as inputs for the color MLP.

We begin to learn the human trajectory modeling at $3k$ iterations, then after $7k$ iterations, we start learning the mask for training stabilization. 
Before $7k$ iterations, we use the averaged blurred rendered result as input of the loss function.
$f_{\theta_{\text{nr}}}$ is an MLP with $3$ hidden layers of $128$ dimensions, and $f_{\theta_{\text{rigid}}}$ is an MLP with $4$ hidden layers of $128$ dimensions as previous works. The color MLP consists of a single $64$-dimensional hidden layer, which has been proven effective~\cite{qian20243dgs}. The mask prediction network $f_{\theta_{\text{fuse}}}$ is an MLP with $4$ hidden layers of $64$ dimensions.

\section{Experiments}
\label{sec:exp}
In this section, we first show how we capture and synthesize the two motion blur human avatar datasets, and we compare with state-of-the-art methods~\cite{weng2022humannerf,wang2022arah,hu2024gauhuman,wen2024gomavatar,qian20243dgs,lei2024gart} quantitatively and qualitatively under two settings, with the original blurred inputs and with the inputs pre-deblurred by state-of-the-art method~\cite{pan2023deep}.
Finally, we conduct an ablation study on each component of the proposed model.
\begin{figure*}[!h]
    \centering    
    \includegraphics[width=\linewidth]{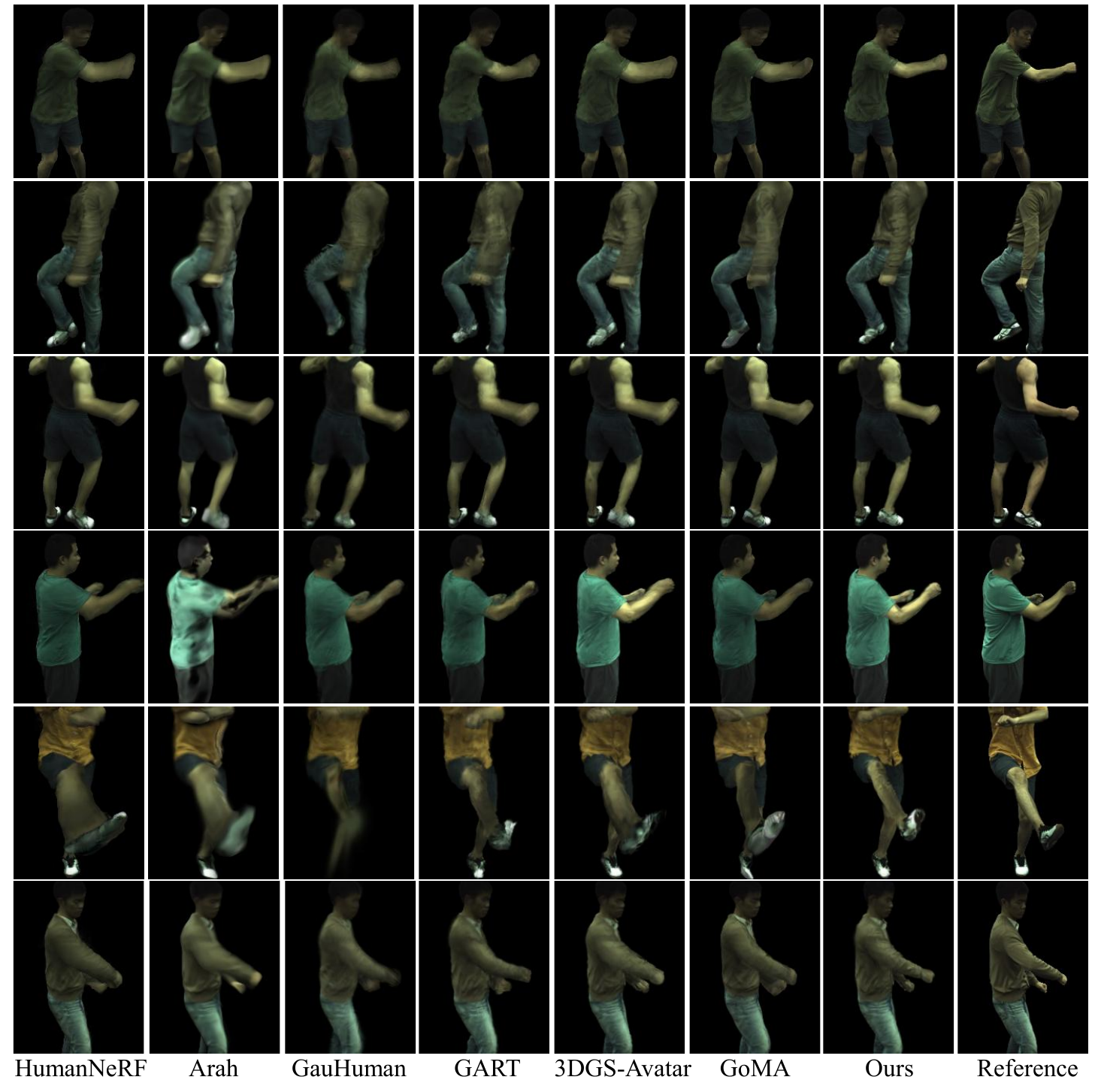}
    \caption{\textbf{Qualitative results on ZJU-MoCap-Blur.} Compared with existing human avatar methods, our method generates sharper novel views and preserves more details with original motion-blurred inputs. 
    }
    \label{fig:baseline}
\end{figure*}
\begin{figure*}[!h]
    \centering    
    \includegraphics[width=0.95\linewidth]{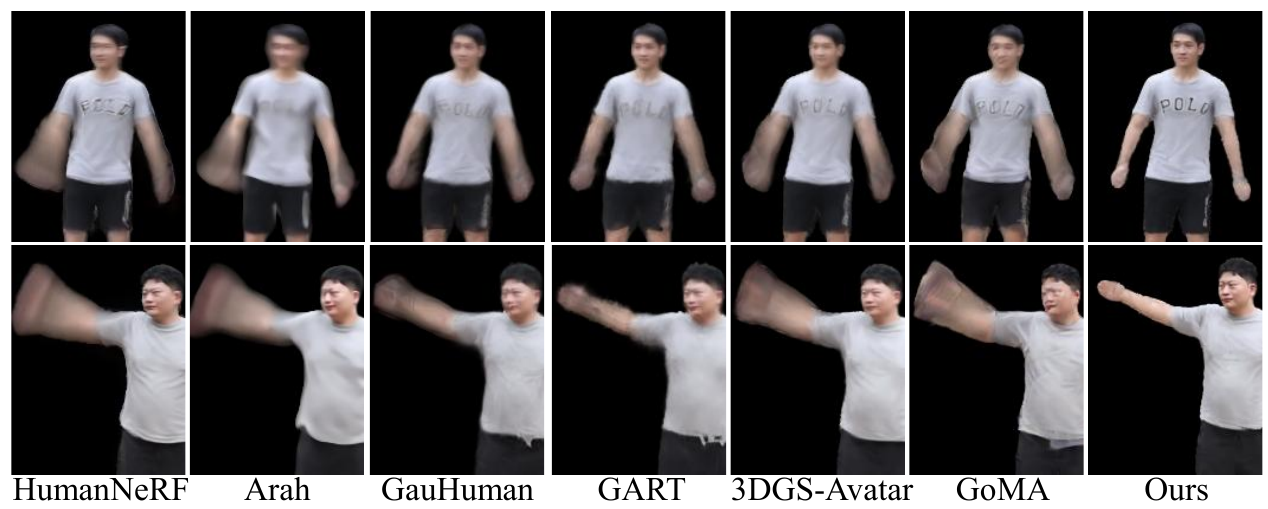}
    \caption{\textbf{Qualitative results on Real-Human-Blur dataset.} Our method generates sharper novel poses that are more faithful and preserve more details. 
    }
    \label{fig:baseline_real}
\end{figure*}

\begin{figure*}[!h]
    \centering    
    \includegraphics[width=1.0\linewidth]{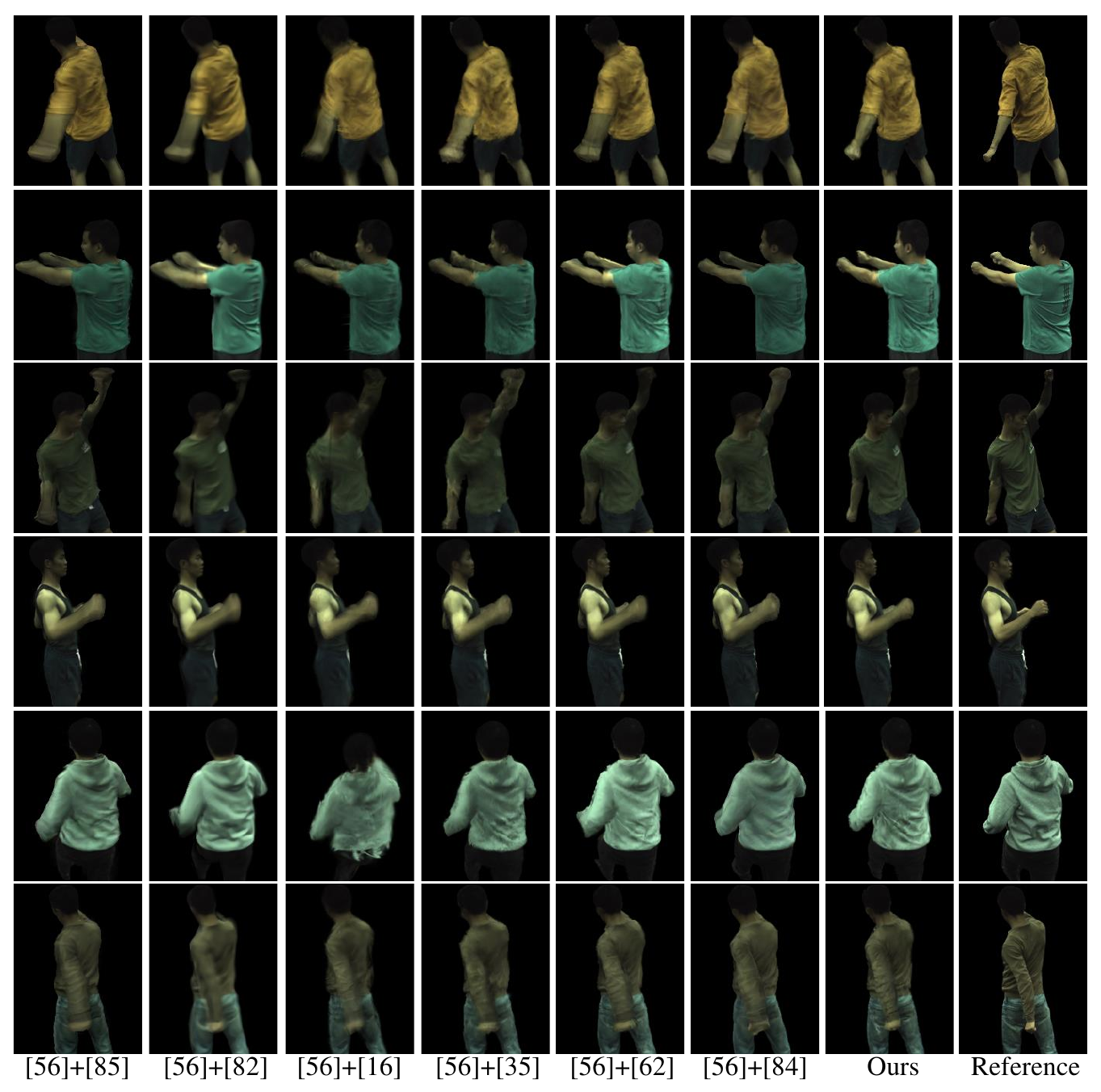}
    \caption{\textbf{Qualitative results on ZJU-MoCap-Blur (pre-deblurred).} The input sequences are deblurred by state-of-the-art video deblurring method~\cite{pan2023deep}. Compared with existing human avatar methods, our method generates sharper novel views and preserves more details with original motion-blurred inputs.
    }
    \label{fig:deblur_supp}
\end{figure*}
\begin{table*}[htp]
	\centering
	\caption{\textbf{Quantitative results on ZJU-MoCap-Blur.} The input sequence is the motion-blurred images.
	The best performance is in \textbf{boldface}, and the second best is \underline{underlined}. We present the training time as GPU and the rendering speed as FPS.} 
	\resizebox{1.0\linewidth}{!}
	{
    \setlength{\tabcolsep}{3pt}
	\renewcommand\arraystretch{1.0}
	\begin{NiceTabular}{lcc|ccc|ccc|ccc|ccc|ccc|ccc}
		\toprule
		\multicolumn{1}{l}{\multirow{2}{*}[-0.5ex]{Method}} 
  &{\multirow{2}{*}[-0.5ex]{GPU}} & {\multirow{2}{*}[-0.5ex]{FPS}} &
  \multicolumn{3}{c}{377} & \multicolumn{3}{c}{386}& \multicolumn{3}{c}{387}& \multicolumn{3}{c}{392}&\multicolumn{3}{c}{393}&\multicolumn{3}{c}{394} \\
		\cmidrule{4-21}
		~&~ & ~ &  PSNR$\uparrow$ &  SSIM$\uparrow$&LPIPS$\downarrow$ & PSNR$\uparrow$ & SSIM$\uparrow$ & LPIPS$\downarrow$ &PSNR$\uparrow$ &  SSIM$\uparrow$ & LPIPS$\downarrow$ &PSNR$\uparrow$ &  SSIM$\uparrow$&LPIPS$\downarrow$&PSNR$\uparrow$ &  SSIM$\uparrow$&LPIPS$\downarrow$&PSNR$\uparrow$ &  SSIM$\uparrow$&LPIPS$\downarrow$ \\
        \midrule
		\midrule
		HumanNeRF~\cite{weng2022humannerf} & 4d & 0.3&29.30 & 0.9634&32.17 & 33.36& 0.9639&31.09&28.15&0.9490&42.86&30.35&0.9520&43.86&27.76&0.9409&50.38&29.05&0.9484&40.44\\
		Arah~\cite{wang2022arah}& 3d & 0.2 & 29.20& 0.9582&48.90& 31.83 &0.9625&40.64 &26.82& 0.9447&60.18& 28.74&0.9464&65.65&27.44&0.9399&70.94&29.43&0.9488&59.13\\
		GauHuman~\cite{hu2024gauhuman}& 4m&189 &29.42&0.9617&33.42&33.46& 0.9660&36.20 & 27.89&0.9420&53.50&28.37 &0.9451&59.57& 27.46 & 0.9377& 63.08&29.17&0.9458&47.71\\
		GART~\cite{lei2024gart}& 4m &142& \underline{30.29}&\textbf{0.9771} &  \underline{28.43}& 33.50 & 0.9659& 34.30& \textbf{29.36}&{0.9618}&44.67&29.80&0.9673&44.26&28.24&0.9603&49.03&29.45&\textbf{0.9652}&43.39\\
		3DGS-Avatar~\cite{qian20243dgs}& 20m&50&29.88 & 0.9733 &  29.38& \underline{33.68}&\underline{0.9764} & 30.44& 28.34&
\underline{0.9621}&41.77&\underline{30.40}&\underline{0.9682}&\underline{41.06}&\underline{28.32}&\underline{0.9604}&\underline{44.58}&\underline{29.74}&0.9633&40.14\\
		GoMA~\cite{wen2024gomavatar}& 1d &44& 29.35 &  0.9722& 30.57 &33.49 & 0.9762&\underline{29.99}&28.10&0.9614& \underline{41.31}&29.86&0.9533&41.72&27.77&0.9437&46.43&29.38&0.9490&\underline{39.30}\\

		Ours& 40m&50&\textbf{30.36} & \underline{0.9765}&\textbf{26.69} & \textbf{33.75}& \textbf{0.9770} &\textbf{29.37}& \underline{28.61}& \textbf{0.9629} & \textbf{40.39}& \textbf{30.45}&\textbf{0.9697}&\textbf{38.49}& \textbf{28.43}&\textbf{0.9617}&\textbf{42.17}&\textbf{29.86}&\underline{0.9646}&\textbf{37.79}\\
    \bottomrule
	\end{NiceTabular}
	}
	\label{tab:baseline}
\end{table*}

\begin{figure*}[!h]
    \centering    
    \includegraphics[width=\linewidth]{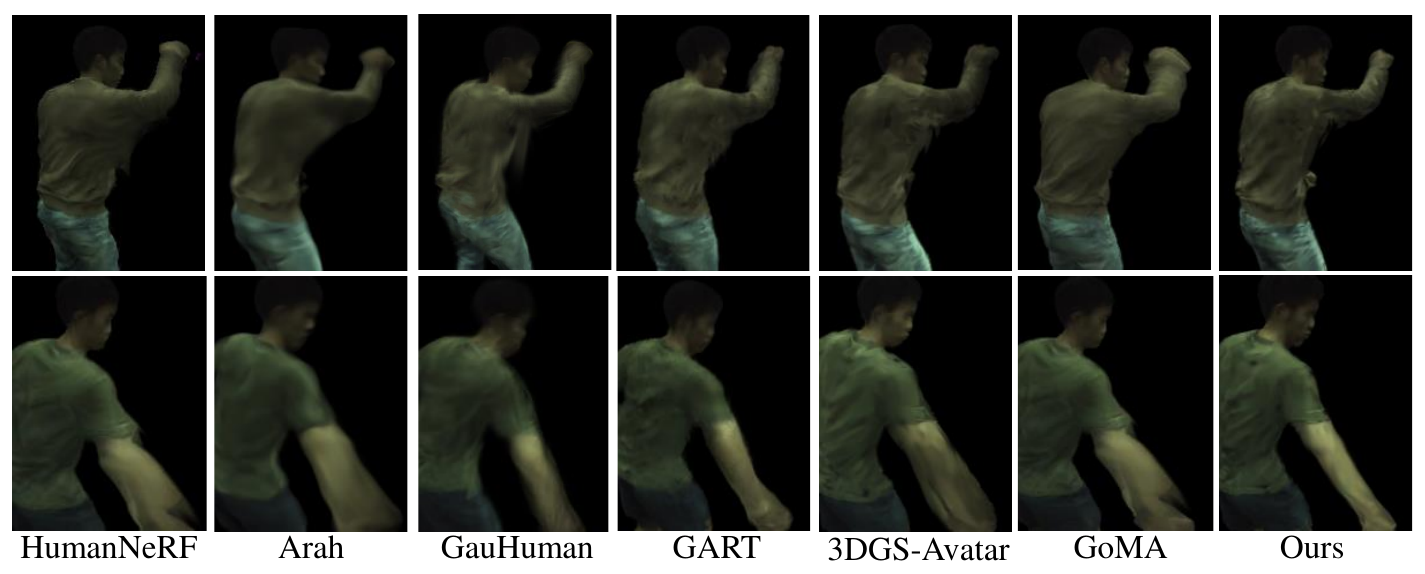}
    \caption{\textbf{The qualitative comparisons of out-of-distribution pose animation on ZJU-MoCap-Blur.} 
    Our method produces fewer artifacts than baselines, demonstrating good generalization to unseen poses.
    }
    \label{fig:ood}
\end{figure*}
\begin{table*}[htp]
	\centering
	\caption{\textbf{Quantitative results on ZJU-MoCap-Blur (pre-deblurred).} The input sequences are pre-deblurred by state-of-the-art video deblurring method~\cite{pan2023deep}.
	The best performance is in \textbf{boldface}, and the second best is \underline{underlined}. We present the training time as GPU and the rendering speed as FPS.} 
	\resizebox{1.0\linewidth}{!}
	{
    \setlength{\tabcolsep}{3pt}
	\renewcommand\arraystretch{1.0}
	\begin{NiceTabular}{lcc|ccc|ccc|ccc|ccc|ccc|ccc}
		\toprule
		\multicolumn{1}{l}{\multirow{2}{*}[-0.5ex]{Method}} 
  &{\multirow{2}{*}[-0.5ex]{GPU}} & {\multirow{2}{*}[-0.5ex]{FPS}} &
  \multicolumn{3}{c}{377} & \multicolumn{3}{c}{386}& \multicolumn{3}{c}{387}& \multicolumn{3}{c}{392}&\multicolumn{3}{c}{393}&\multicolumn{3}{c}{394} \\
		\cmidrule{4-21}
		~&~ & ~ &  PSNR$\uparrow$ &  SSIM$\uparrow$&LPIPS$\downarrow$ & PSNR$\uparrow$ & SSIM$\uparrow$ & LPIPS$\downarrow$ &PSNR$\uparrow$ &  SSIM$\uparrow$ & LPIPS$\downarrow$ &PSNR$\uparrow$ &  SSIM$\uparrow$&LPIPS$\downarrow$&PSNR$\uparrow$ &  SSIM$\uparrow$&LPIPS$\downarrow$&PSNR$\uparrow$ &  SSIM$\uparrow$&LPIPS$\downarrow$ \\
        \midrule
		\midrule
		~\cite{pan2023deep}+~\cite{weng2022humannerf}  & 4d & 0.3&29.32&0.9623  & 31.93&33.43&0.9646&30.60 & 28.10& 0.9468& 44.33&\underline{30.30}&0.9520&43.25&27.80&0.9413&49.56&28.92&0.9481&40.12\\
		~\cite{pan2023deep}+~\cite{wang2022arah}& 3d & 0.2 &28.97&  0.9560&48.91&32.15&0.9663&38.62& 27.32&0.9477&55.92&29.76 & 0.9516& 59.08&\underline{28.36}&0.9449&64.91&29.40&0.9485&59.84\\
		~\cite{pan2023deep}+~\cite{hu2024gauhuman}& 4m&189&29.20 & 0.9625&34.39 & 33.49&0.9659 & 34.30& 27.88&0.9423&53.44&28.65&0.9464&57.99&27.72&0.9371&61.81&29.11&0.9456&47.68\\
		~\cite{pan2023deep}+~\cite{lei2024gart}& 4m&142&\underline{30.27} & \textbf{0.9770}&\underline{28.31}&\textbf{33.86}&\textbf{0.9770} & 33.22&\textbf{29.34} &  \textbf{0.9658}&44.48&  29.77&0.9671&44.21&28.22&\underline{0.9602}&49.02&29.41&\underline{0.9651}&43.14\\
		~\cite{pan2023deep}+~\cite{qian20243dgs}&20m&50& {29.69}& 0.9732 &  29.26&33.66&\underline{0.9765}& \underline{29.92} &28.38& {0.9622}& 41.48&{30.29}&\underline{0.9678}&\underline{41.10}&28.30&0.9596&\underline{44.94}&\underline{29.67}&0.9634&39.70\\
		~\cite{pan2023deep}+~\cite{wen2024gomavatar}& 1d&44&29.20 & 0.9716 & 30.93& 33.47&0.9761&29.98&28.02&0.9613&\underline{41.11} & 29.90& 0.9532&41.81&27.84&0.9440&46.07&29.28&0.9485&\underline{39.52}\\
        
		Ours& 40m&50&\textbf{30.36} & \underline{0.9765}&\textbf{26.69} & \underline{33.75}& \textbf{0.9770} &\textbf{29.37}& \underline{28.61}& \underline{0.9629} & \textbf{40.39}& \textbf{30.45}&\textbf{0.9697}&\textbf{38.49}& \textbf{28.43}&\textbf{0.9617}&\textbf{42.17}&\textbf{29.86}&\textbf{0.9646}&\textbf{37.79}\\
    \bottomrule
	\end{NiceTabular}
	}
	\label{tab:baseline_deblur}
    \vspace{-6pt}
\end{table*}
\subsection{Datasets}
There are no available datasets for motion-blurred inputs in animatable human avatar tasks, so we curated two datasets: (1) Synthesized dataset from ZJU-MoCap~\cite{peng2021neural} and (2) Real blur dataset from our capturing and internet videos.

\noindent\textbf{ZJU-MoCap-Blur.} This is the main dataset for quantitative evaluation. We pick six sequences (377, 386, 387, 392, 393, 394) from the ZJU-MoCap dataset and follow the training/test split of HumanNeRF~\cite{weng2022humannerf}. We synthesize motion blur as the pipeline of current benchmark datasets on motion deblurring~\cite{nah2017deep,zhou2019davanet,nah2019ntire}. 
ZJU-MoCap is shot at a frame rate of 60 fps. To synthesize realistic motion blur without artifacts like previous work~\cite{nah2017deep}, we increase the video frame
rate to 480 fps using a state-of-the-art frame interpolation method~\cite{zhang2023extracting}.
Then we average this sharp high frame rate
successive frames to generate a blurry image to approximate a long
exposure time, and the generated frames are temporally centered
on a real-captured ground truth frame from original ZJU-MoCap. 
We apply varying blur sizes for the dataset, so the dataset consists of scenes with small blur(17 frames to synthesize one frame) for sequences 377 and 392, medium blur (33 frames) for 393 and 394, and large blur (49 frames) for sequences 386 and 387.
To make the dataset more realistic, we use EasyMocap~\cite{dong2020motion,peng2021neural} to re-calculate the human poses and the human masks of the synthesized blurred image sequences.
We train and evaluate the Real-Human-Blur dataset at the resolution of $540\times540$, $960\times540$, and $360\times640$, depending on the original resolution of the captured video.
Following previous works~\cite{qian20243dgs,weng2022humannerf}, we conduct quantitative evaluations on novel view synthesis and show qualitative results for animation on out-of-distribution poses. 
LPIPS in all the tables are scaled up by 1000.

\noindent\textbf{Real-Human-Blur.}
Given the absence of publicly available, real-world datasets that are specifically tailored for tackling the challenge of motion blur in human avatar modeling, we have curated a dataset consisting of monocular motion-blurred videos. 
The videos were recorded using a high-resolution DSLR camera under varying lighting and background settings, allowing for rich visual details and realistic blur patterns caused by human motion. 
We use SPIN~\cite{kolotouros2019learning} to obtain approximate body poses and employ SAM~\cite{ravi2024sam2} for segmenting the foreground human. 
Since there is no accurate sharp ground truth for these real captures, we use this dataset solely for qualitative comparisons on novel pose synthesis.

\subsection{Baseline Comparisons}
The baselines include two types: 1) state-of-the-art human avatar reconstruction methods, e.g., NeRF-based methods HumanNeRF~\cite{weng2022humannerf} and ARAH~\cite{wang2022arah}, 3DGS-based methods GauHuman~\cite{hu2024gauhuman}, GoMAvatar~\cite{wen2024gomavatar}, 3DGS-Avatar~\cite{qian20243dgs} and GART~\cite{lei2024gart} 
and 2) an image-space baseline that uses pre-trained video deblurring~\cite{pan2023deep} for preprocessing and then trains the human avatar baselines with the deblurred inputs.
These baselines are compared under the monocular setup on ZJU-MoCap-Blur and Real-Human-Blur. 
All experiments are conducted on an NVIDIA RTX 3090 GPU. 

\subsection{Qualitative Results}

In \cref{fig:baseline}, we present a comparative evaluation of our method against several state-of-the-art human avatar modeling approaches on the ZJU-MoCap-Blur dataset. As shown in the figure, current approaches struggle to reconstruct fine-grained details from motion-blurred inputs.
In \cref{fig:baseline_real}, we show the novel pose synthesis results on the  Real-Human-Blur dataset. 
One can observe that current human avatar modeling methods cannot recover sharp details from motion blur, and our method outperforms these baselines in handling motion blur and generating sharp novel views and poses. 
We visualize the out-of-distribution pose animation on ZJU-MoCap-Blur with pose sequences from AMASS~\cite{mahmood2019amass} and AIST++~\cite{li2021ai} in \cref{fig:ood}. This demonstrates our model's generalization ability to extreme out-of-distribution poses.
We use an off-the-shelf~\cite{kolotouros2019learning} pose estimator for the in-the-wild Real-Human-Blur dataset, yet we still achieve satisfying results.
We also present the qualitative comparisons on the baseline models with video deblurring preprocessing, as shown in \cref{fig:deblur_supp}.

\subsection{Quantitative Results}
The quantitative results on ZJU-MoCap-Blur are reported in \cref{tab:baseline} and \cref{tab:baseline_deblur}. 
The baseline model in \cref{tab:baseline} is trained on original motion-blurred inputs, and the inputs in \cref{tab:baseline_deblur} are pre-processed by video deblurring. 
Overall, our proposed approach performs better on PSNR and SSIM, and outperforms all the baselines on LPIPS, which is more informative in a monocular setting~\cite{qian20243dgs}. 
Video deblurring preprocessing improves the baselines only by a small margin, because of its focus on blur from camera motion and this 2D image-space paradigm lacks 3D consistency.
Our method is capable of fast training and renders at a real-time rendering frame rate. Human motion trajectory modeling and pose-dependent fusion do not increase too much training time and do not add any rendering cost. 
Although Gauhuman~\cite{hu2024gauhuman} performs well on the original ZJU-MoCap dataset, discarding non-rigid deformation and implementing a new optimization pipeline for fast convergence result in its poor performance on the motion-blurred inputs.
We ensure all methods are evaluated under the same version of SSIM and LPIPS calculators because different versions lead to numerical differences.
GauHuman~\cite{hu2024gauhuman} is trained for 7k iterations instead of 3k as original paper to get the best performance.

\subsection{Ablation Study}

\begin{figure}[!h]
    \centering    
    \includegraphics[width=1.0\linewidth]{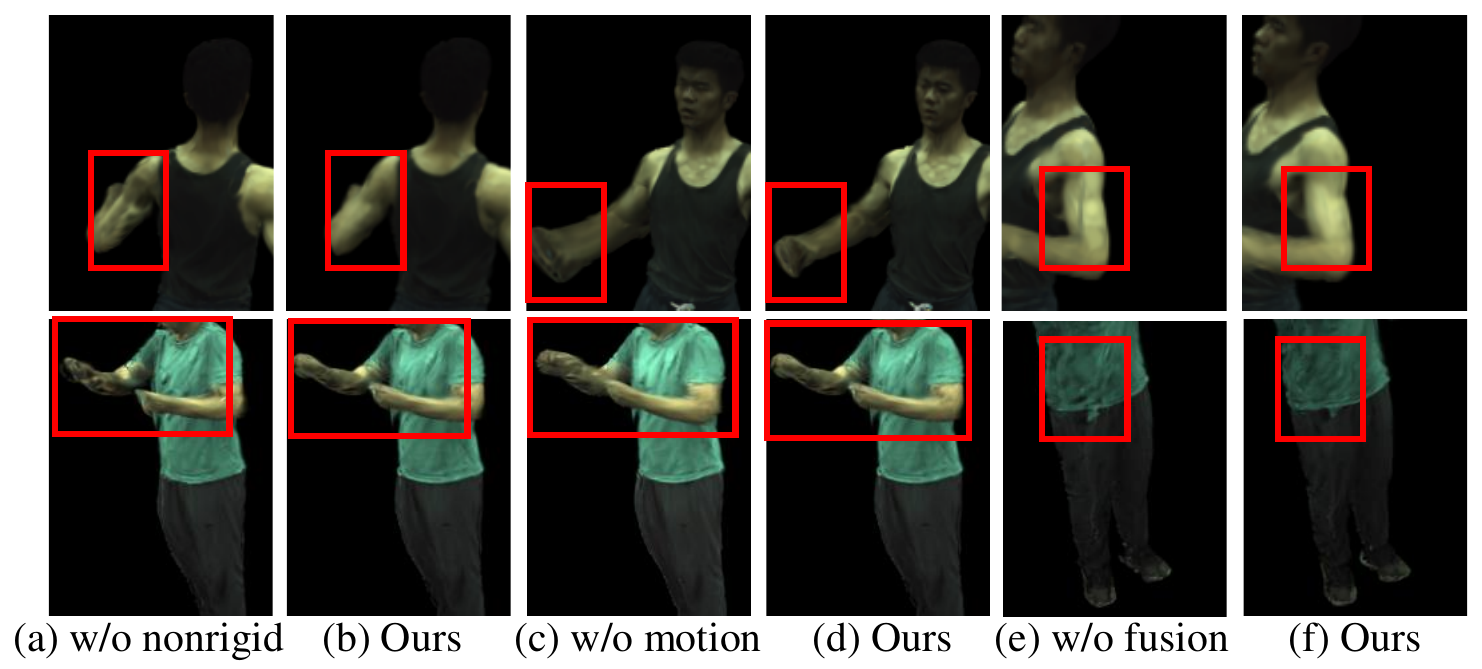}
    \caption{\textbf{Ablation Study on non-rigid deformation, motion trajectory modeling, and pose-dependent fusion.} 
    Implementing these modules preserves more details and alleviates avatars' motion-related artifacts.
    }
    \label{fig:ablation}
    \vspace{-8pt}
\end{figure}

Our proposed method addresses the motion blur by predicting human pose sequences from human movements and incorporating the motion blur formation into 3DGS training by pose-dependent fusion. 
Therefore we conduct ablation studies on the human motion trajectory modeling module and the fusion module. 
As shown in \cref{fig:ablation} and \cref{tab:ablation study}, our framework works best when all the components are applied. 
We also conduct experiments on the number of virtual human poses in the exposure time in \cref{tab:virtual number}, the results show that increasing virtual poses is not always optimal if we aim to balance efficiency and quality.

The results of our ablation study on trajectory representations are presented in \cref{tab:trajectory}. 
We conduct on two settings: (1) optimizing $\theta_\text{start}$ and $\theta_\text{end}$ and interpolate the human poses by linear interpolation, and (2) utilizing a high-order cubic B-spline that jointly optimizes four control knots to depict human motions.
The results demonstrate that Spherical Linear Interpolation adequately represents the human motion trajectory.

We also explore whether adding a learnable interpolation will further improve the performance in \cref{tab:trajectory_learn}. The results show that the current interpolation method is adequate for human motion trajectory modeling.

\begin{table}[t]
\caption{\textbf{Ablation study on the effectiveness of our proposed modules.} We evaluate the six scenes of ZJU-MoCap-Blur, and we present the performance on average. }
    \centering
    \setlength{\tabcolsep}{5pt}
    \begin{tabular}{@{}lccc@{}}
        \toprule
        \small Method & \small PSNR$\uparrow$ & \small SSIM$\uparrow$ & \small LPIPS$\downarrow$ \\
        \midrule
        \small w/o non-rigid &\small 29.43&\small 0.9659&\small38.12\\
        \small w/o motion modeling & \small \underline{30.05} & \small{0.9668} & \small37.95\\
         \small w/o fusion & \small29.83 & \small\underline{0.9676} & \small\underline{36.42} \\
        \midrule
         \small Full (Ours) & \small \textbf{30.24} & \small \textbf{0.9684} & \small \textbf{35.82} \\
        \bottomrule
    \end{tabular}\hskip1.2pt
    
    \label{tab:ablation study}
    \vspace{-8pt}
\end{table}

\begin{table}
\caption{\textbf{Ablation study on the number of virtual human poses n.} We evaluate on two scenes. The results indicate that performance does not necessarily improve with increasing number n, but the increasing number will prolong the training time.}
    \centering
    \setlength{\tabcolsep}{3pt}
    \subfloat{
    \begin{tabular}{@{}lccc@{}}
        \toprule
        \small $377$ & \small PSNR$\uparrow$ & \small SSIM$\uparrow$ & \small LPIPS$\downarrow$ \\
        \midrule
        \small 3& \small 30.03 & \small0.9757 & \small27.20 \\
         \small5& \small \textbf{30.36} & \small\textbf{0.9765} & \small\textbf{26.69} \\
         \small7&\small\underline{30.27}& \small\underline{0.9763} &  \small\underline{26.91}\\
        \small9& \small30.20 &\small 0.9759 &\small27.31 \\
        \small13& \small30.18 & \small0.9759 &\small27.12 \\
        \bottomrule
    \end{tabular}}\hskip1.2pt
    \subfloat{
    \begin{tabular}{@{}lccc@{}}
        \toprule
         \small $386$ & \small PSNR$\uparrow$ & \small SSIM$\uparrow$ & \small LPIPS$\downarrow$ \\
        \midrule
         \small3& \small33.30 &\small 0.9764 & \small30.02 \\
        \small5& \small33.75& \small0.9770 & \small\underline{29.37} \\
         \small7& \small\underline{33.78} & \small0.9770&\small29.50\\
         \small9& \small\textbf{33.79} & \small\underline{0.9772} & \small\textbf{29.36} \\
         \small13& \small33.77 & \small\textbf{0.9773} &\small29.48  \\
        \bottomrule
    \end{tabular}}
    \label{tab:virtual number}
\end{table}

\begin{table}[!h]
\caption{\textbf{Ablation study on the trajectories representations.} We evaluate four scenes from ZJU-MoCap-Blur. The best performance of each scene is \textbf{boldfaced} on each row.}
    \centering
    \setlength{\tabcolsep}{2pt}
    \subfloat{
    \begin{tabular}{@{}lccc@{}}
        \toprule
        \small Slerp& \small PSNR$\uparrow$ & \small SSIM$\uparrow$ & \small LPIPS$\downarrow$ \\
        \midrule
        \small377& \small\textbf{30.36}& \small\textbf{0.9765} & \small\textbf{26.69} \\
         \small386&  \small33.75 &\small0.9770&\small\textbf{29.37}\\
         \small387&\small28.61& \small0.9629& \small\textbf{40.39} \\
        \small392& \small\textbf{30.45} & \small\textbf{0.9697} &\small\textbf{38.49}\\
        \bottomrule
    \end{tabular}}\hskip1.2pt
    \subfloat{
    \begin{tabular}{@{}lccc@{}}
        \toprule
         \small Cubic & \small PSNR$\uparrow$ & \small SSIM$\uparrow$ & \small LPIPS$\downarrow$ \\
        \midrule
         \small377& \small30.19 & \small0.9758 & \small27.58 \\
         \small386& \small\textbf{33.86}& \small\textbf{0.9772} & \small29.49 \\
         \small387&\small\textbf{28.66}&\small \textbf{0.9630} & \small40.62 \\
        \small392& \small30.43 & \small0.9694 &\small39.33 \\
        \bottomrule
    \end{tabular}}
    \label{tab:trajectory}
\end{table}
\begin{table}[!h]
\caption{\textbf{Ablation study on the learnable trajectories representations.} We evaluate two scenes from ZJU-MoCap-Blur. The best performance of each scene is \textbf{boldfaced} on each row.}
    \centering
    \setlength{\tabcolsep}{2pt}
    \subfloat{
    \begin{tabular}{@{}lccc@{}}
        \toprule
        \small Slerp& \small PSNR$\uparrow$ & \small SSIM$\uparrow$ & \small LPIPS$\downarrow$ \\
        \midrule
        \small377& \small\textbf{30.36}& \small\textbf{0.9765} & \small\textbf{26.69} \\
         \small386&  \small{33.75} &\small0.9770&\small\textbf{29.37}\\
        \bottomrule
    \end{tabular}}\hskip1.2pt
    \subfloat{
    \begin{tabular}{@{}lccc@{}}
        \toprule
         \small Learn & \small PSNR$\uparrow$ & \small SSIM$\uparrow$ & \small LPIPS$\downarrow$ \\
        \midrule
         \small377& \small30.31 & \small 0.9759 & \small27.81 \\
         \small386& \small\textbf{33.85}& \small \textbf{0.9773}& \small29.51 \\
        \bottomrule
    \end{tabular}}
    \label{tab:trajectory_learn}
\end{table}

\subsection{Computational Efficiency}
In terms of resource requirements, we conduct experiments to calculate the training and inference costs. Our method costs more memory than 3DGS baselines due to the increased complexity of modeling motion blur, but saves more memory than NeRF baselines in \cref{tab:supp_com}. This trade-off highlights the balance our method strikes between performance and computational cost, offering a practical alternative for high-quality human avatar modeling under motion blur.
\begin{table}[ht]
\caption{\textbf{Training and inference efficiency.}}
\label{tab:supp_com}
\centering
\resizebox{1.0\linewidth}{!}{
\begin{tabular}{@{}ccccccccc@{}}
\toprule
\multicolumn{2}{c}{Method} & HumanNeRF &Arah& GauHuman & GART &3DGS-Avatar&GoMA&Ours\\
\midrule
\multicolumn{2}{c}{Training Memory (G)} & 11.3&11.9& 2.0&2.5&6.4&4.4&7.3 \\
\multicolumn{2}{c}{Inference Memory (G)} & 18.7&19.2&0.9& 3.3&3.1&3.2&2.7\\

\bottomrule
\end{tabular}
}

\end{table}

\subsection{Difference With Current Deblurring Methods}
\begin{figure}[!h]
    \centering    
    \includegraphics[width=1.0\linewidth]{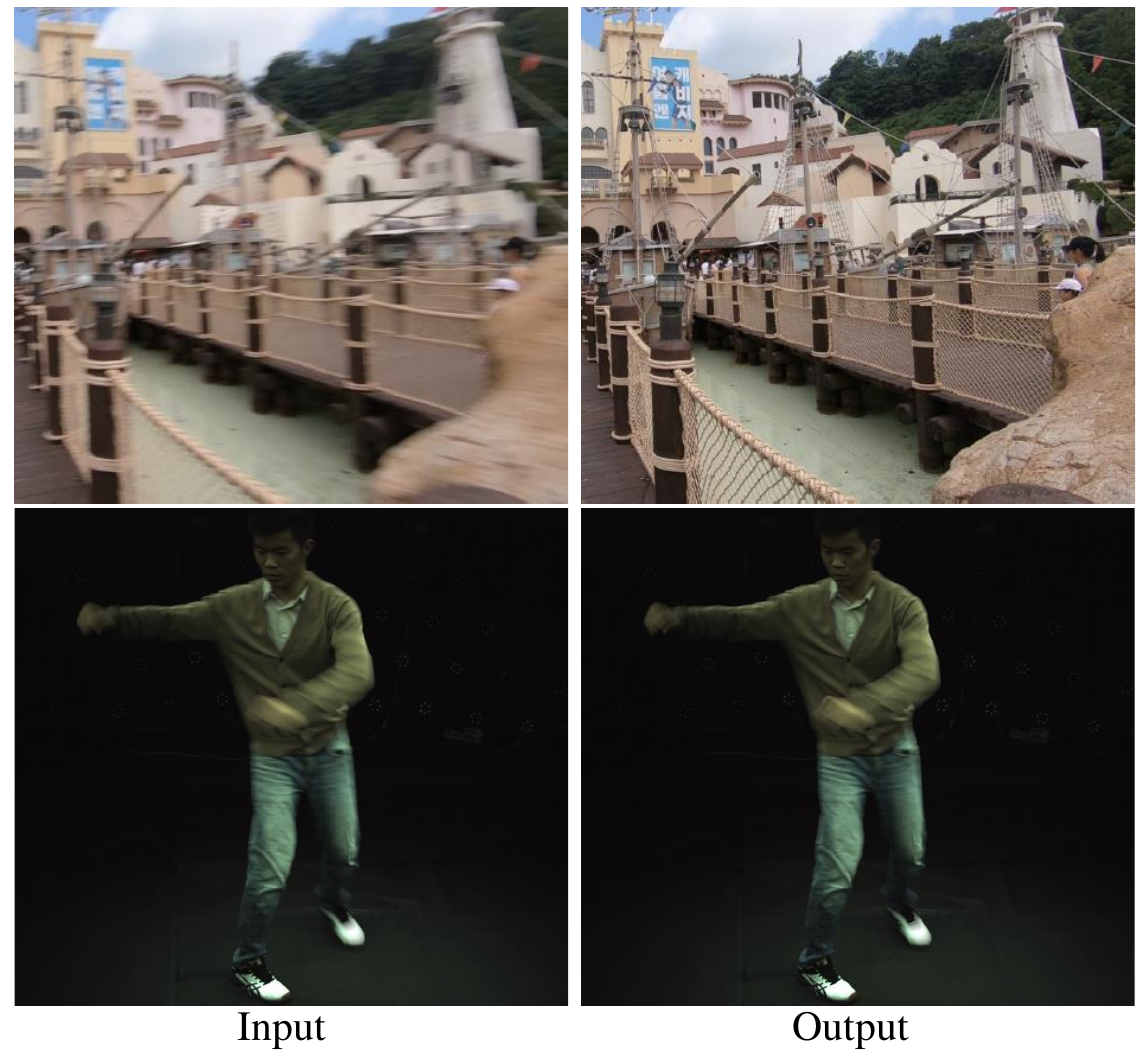}
    \caption{\textbf{Pre-deblurring results of state-of-the-art video deblurring method~\cite{pan2023deep}.} On the first row we show the data where this method performs well. Current video deblurring paradigms are fitting for motion blur from camera shake. However, the motion blur in the human avatar modeling task is primarily from human movements, and current methods perform poorly.
    }
    \label{fig:deblur_example_supp}
\end{figure}

As shown in \cref{tab:baseline_deblur} and \cref{fig:deblur_supp}, the current state-of-the-art video deblurring method~\cite{pan2023deep} only improves the human avatar modeling baselines by a small margin. Therefore, we conduct qualitative experiments solely on the performance of the video deblurring method. As shown in \cref{fig:deblur_example_supp}, the current deblurring pipeline works well on motion blur due to camera movements. However, the pipeline fails to handle motion blur from human movements, because the blur pattern of the input is different.
\section{Conclusion}
We present a novel framework for sharp reconstruction of clothed human avatars from motion-blurred monocular video. 
To tackle motion blur from human movements in video capture, we model the human motion trajectories in 3DGS. Each timestamp is regarded as a sequence of sharp images captured within exposure time, and a sequence of human poses is predicted based on the input parameters. As human movements rarely involve the whole body, we predict masks to indicate the degradation regions within an image, enabling effective training on both blurred regions and sharp regions. 
Extensive experiments show that our real-time rendering method produces high-quality sharp avatars compared to state-of-the-art works. 


\noindent\textbf{Limitations.} 
(1) We rely solely on a single input pose for trajectory modeling. A potential future direction is to explore inter-frame relationships to improve pose trajectory representation. 
(2) The proposed method does not reconstruct the accurate geometry of the avatar, and a potential direction is to extract smooth geometry from the 3DGS human avatar model by incorporating mesh or regularizing normal map. (3) Like previous works, our method may produce poorly with high-frequency details, e.g., complex local cloth, and the refinement for these areas is yet to be explored.




{
    \small
    \bibliographystyle{IEEEtran}
    \bibliography{refs}
}

\end{document}